\documentclass{article}

\usepackage{amsmath,amsfonts,amssymb,times,graphicx,natbib,algorithm,algorithmic,hyperref}

\usepackage[accepted]{hill2019}

\usepackage{microtype}
\usepackage{graphicx}
\usepackage{booktabs} 
\usepackage{amssymb}
\usepackage{subcaption}
\usepackage{color}

\icmltitlerunning{Extracting Interpretable Concept-Based Decision Trees from CNNs}

\begin{document}

\twocolumn[
\icmltitle{Extracting Interpretable Concept-Based Decision Trees from CNNs}


\icmlsetsymbol{equal}{*}

\begin{icmlauthorlist}
\icmlauthor{Conner Chyung}{usc}
\icmlauthor{Michael Tsang}{usc}
\icmlauthor{Yan Liu}{usc}
\end{icmlauthorlist}

\icmlaffiliation{usc}{University of Southern California, Los Angeles, CA, USA}

\icmlcorrespondingauthor{Conner Chyung}{cchyung@usc.edu}

\icmlkeywords{interpretability, decision trees, concepts}

\vskip 0.3in
]


\printAffiliationsAndNotice{}
\begin{abstract}
In an attempt to gather a deeper understanding of how convolutional neural networks (CNNs)
reason about human-understandable concepts,we present a method to infer labeled concept data from hidden layer activations and interpret the concepts through a shallow decision tree. 

The decision tree can provide information about which concepts a model deems important, as well as provide an understanding how the concepts interact with each other.  
Experiments demonstrate that the extracted decision tree is capable of accurately representing the original CNN's classifications at low tree depths, thus encouraging human-in-the-loop understanding of discriminative concepts.
\end{abstract}

\section{Introduction}
It is generally understood that Convolutional Neural Networks learn abstract, semantic concepts, but there is still an ongoing question about how the model uses these concepts and how they inform the model's prediction. Motivated by a desire to explain why a CNN makes the decisions it does in a human-interpretable manner, we propose a method that formulates a global interpretation of the semantic concepts the model is reasoning about~\citep{ribeiro2016should, kim2017interpretability} using a shallow decision tree based on concept data extracted from activations at the hidden layer.  This method is both efficient and portable. It does not require retraining of any existing model one wishes to test.

Because CNNs are largely black box systems, this global interpretation can be valuable as it grants a general understanding of how the model is behaving and provides a logical explanation for the decisions that the model makes.  This kind of interpretation can increase confidence and trust in the model if it is found that it is making decisions that seem reasonable to humans.

Understanding how semantic concepts inform the decision the model is making can also be used to highlight potential unwanted bias learned by the model based on the most discriminating concepts learned by the decision tree.  

Using a densely labeled image data set to probe the network, we show that for a classification problem with few classes, a shallow, interpretable decision tree can be learned that is nearly as accurate as the original model.  We also demonstrate that the shallow decision tree learned performs comparably well to deeper, but less interpretable decision trees. 

\section{Related Works}
\textbf{Concepts:} 
Much work has been done on the extraction of concepts learned in the CNN hidden layer. \citet{DBLP:journals/corr/abs-1801-03454} showed that combinations of filters are needed to encode a specific concept and showed how concept classifiers can be trained to recognize the presence of concepts in activations.  \citet{kim2017interpretability} presents a method which gives the ability to extract Concept Activation Vectors and test how sensitive a certain prediction is to a specific concept.  

\textbf{Decision Trees and Neural Networks:}
\citet{balestriero2017neural} presents a hybrid architecture of a decision tree and a neural network which is able to sometimes achieve an accuracy better than its neural network counterparts for specific problems. \citet{DBLP:journals/corr/abs-1711-09784} shows how filter activations themselves can be used to train a decision tree, but the nodes of those trees don't necessarily communicate semantic meaning about what the model is deciding on.  \citet{zhang2018interpreting} is able to learn a decision tree based on semantic meaning.  However their method requires a retraining of the entire network to get each filter to recognize a specific concept before being able to train a decision tree.

Our method is unique in that it provides a global and interpretable explanation of the CNN using a decision tree that shows how concepts interact without having to retrain the network being probed.

\section{Methods}
\subsection{Probing the CNN to Train Concept Classifiers}
Consider a densely labeled image dataset 
$\mathcal{D}=\{(\mathbf{x}^{(1)}, \mathbf{y}^{(1)}), (\mathbf{x}^{(2)}, \mathbf{y}^{(2)}), \dots, (\mathbf{x}^{(n)}, \mathbf{y}^{(n)})\}$ 
with $n$ data points labeled according to a set of concepts $\mathcal{C}$. 
$\mathbf{x} \in \mathbb{R}^d$ is an image with dimensionality $d$ and $\mathbf{y} \in \{0,1\}^{|\mathcal{C}|}$ is a vector of binary variables indicating the presence of a concept in $\mathbf{x}$.

Given a pretrained image classification model $m$, for each image $\mathbf{x} \in \mathcal{D}$, the hidden layer activations  $\mathbf{v}=m_l(\mathbf{x})$ at layer $l$ are extracted and stored alongside the corresponding concept labels $\mathbf{y}$. 

For each concept $c \in \mathcal{C}$, we train a binary linear classifier $f_c$ on a dataset $\mathcal{G}_c$ which is based on dataset $\mathcal{D}$. We define $\mathcal{G}_c = \mathcal{G}_c^+ \cup \mathcal{G}_c^-$ where $\mathcal{G}_c^+=\{(m_l(\mathbf{x}^{(1)}), y_c^{(1)}), \dots, (m_l(\mathbf{x}^{(n)}),{y_c^{(n)})}|_{y_c=1}\}$ and $\mathcal{G}_c^-=\{(m_l(\mathbf{x}^{(1)}), y_c^{(1)}), \dots, (m_l(\mathbf{x}^{(n)}), y_c^{(n)})|_{y_c=0}\}$.

In order to balance the data used to train $f_c$, $\mathcal{G}_c^-$ is taken as a randomly sampled set of negative examples such that $|\mathcal{G}_c^+| = |\mathcal{G}_c^-|$

\textit{\textbf{Note:} As is common in the image classification domain, sometimes size of the hidden layer activation vectors is too large.  To reduce the dimensionality of the concept classification problem, principle component analysis is applied to transform the activations to a reasonable width to train $f$.  Additionally, spatial averaging is also applied if necessary.}

\subsection{Extracting Concept Data}
Consider an image classification problem with dataset $\mathcal{A}=\{(\mathbf{x}^{(1)}, y^{(1)}), (\mathbf{x}^{(2)}, y^{(2)}), \dots, (\mathbf{x}^{(n')}, y^{(n')})\}$ with $n'$ images and $y \in \{1,2,\dots,\gamma\}$, where $\gamma$ is the number of classes.  This time, $\mathbf{x} \in \mathbb{R}^d$ remains an image with dimensionality $d$, and now $y$ is the class label for $\mathbf{x}$.

For each image, $\mathbf{x} \in \mathcal{A}$, hidden layer activations  $\mathbf{v}=m_l(\mathbf{x})$ for the same layer $l$ are extracted from the network. If PCA and/or spatial averaging was applied on the activations during the probing step, the same transformations are applied to $\mathbf{v}$ to achieve the same input dimensionality for $f_c$. 

We use the concept classifiers to make a binary prediction for each $r_c=f_c(\mathbf{v}), r_c \in \{0,1\}$, to create a binary vector $\mathbf{v'}=(r_1, r_2, \dots, r_{|\mathcal{C}|})$, representing whether or not each concept was present in $\mathbf{x}$.

The class prediction $\hat{y}=m(\mathbf{x})$ is also recorded to be used as the target output for training the decision tree.

\begin{figure}[ht]
\begin{center}
\centerline{\includegraphics[width=0.95\columnwidth]{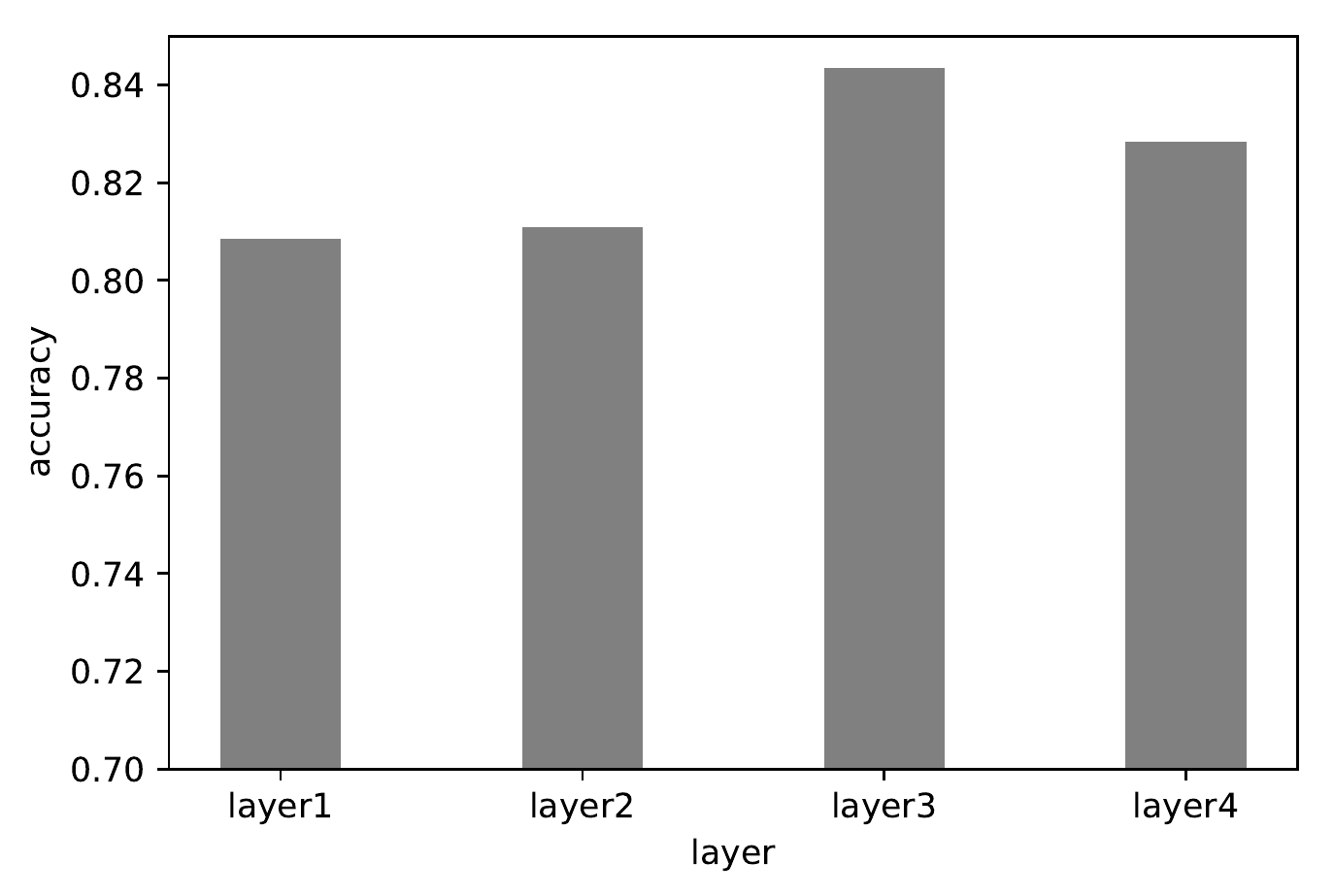}}
\caption{Average accuracy of all concept classifiers trained for each layer.  Concept classifiers for layer 3 have a relatively higher accuracy.}
\label{fig:avg_clf_accuracy_per_layer}
\end{center}
\vspace{-0.2in}
\end{figure}

\subsection{Training the Concept Decision Tree}
The concept vector $\mathbf{v'}$ predicted for each image from the classification problem, $\mathbf{x} \in \mathcal{A}$, as well as the corresponding prediction of $m(\mathbf{x})$ is used to train a decision tree.  We use the default decision tree algorithm from \textit{scikit-learn} \citep{scikit-learn}.  The accuracy of the tree is calculated based on the prediction of $m$ instead of the ground truth label to get the validation accuracy of the decision tree with respect to the representation learned by $m$.

\textit{\textbf{Note:} At the time of writing, the algorithm that \textit{scikit-learn} used for decision tree training was an optimized version of CART (Classification and Regression Trees)}

\section{Results}
\subsection{Data}
For the densely labeled image set used to extract concepts learned by the network we use BRODEN from \citet{DBLP:journals/corr/BauZKOT17}.  The BRODEN dataset is a collection of over 60,000 images with segmentations of concepts belonging to a number of abstract categories including materials, colors, and scenes.  

BRODEN contains over a 1189 different concept labels belonging to different broader categories such as \textit{material, scene} and \textit{color}.  However, some concepts in the dataset have much fewer labeled examples than the others.  Concepts with less than $1000$ examples were unused, leaving around 200 potential concept labels.  

Because a majority of concepts were labeled at the pixel level, additional pre-processing was required to find every concept present in the image overall.  Each image was iterated over and tagged for a specific concept if there were pixels in the image that were labeled with that same concept.

To extract concept decisions from a pre-trained model, the Natural Images dataset from Kaggle was used~\citep{roy2018effects}.  The Natural Images dataset consists of 6899 images with 8 distinct classes (airplane, car, cat, dog, flower, fruit, motorbike, person).

A subset of the Natural Images dataset, \textit{Mini Natural Images}, with only 4 of the 8 classes is also used in a separate experiment (flower, dog, car, person).  This version of the dataset consisted of 3499 labeled examples.

\subsection{Experiment Setup}
We probe Resnet50 pre-trained on the Imagenet dataset.  ~\citep{he2016deep}.

The BRODEN dataset is used as the densely labeled image set to train concept classifiers for the network. Activations from all 4 \textit{major layers} of Resnet50 are extracted.  \textit{Major layers} refer to the \textit{conv2\_x}, \textit{conv3\_x}, \textit{conv4\_x}, and \textit{conv5\_x} blocksections of sublayers of Resnet50.  Spatial averaging and PCA are applied to lower the dimensionality of the activations.  Additionally, in order to ensure the quality of the concept data produced, concept classifiers with validation accuracy scores below $\lambda=0.75$ were discarded.

To create a toy image classification scenario, we retrain the classification layer Resnet50 on the Natural Images dataset.  All layers before the classification layer are frozen to maintain the representation that was that was learned from ImageNet.

\begin{figure}[t]
\begin{center}
\centerline{\includegraphics[width=0.95\columnwidth]{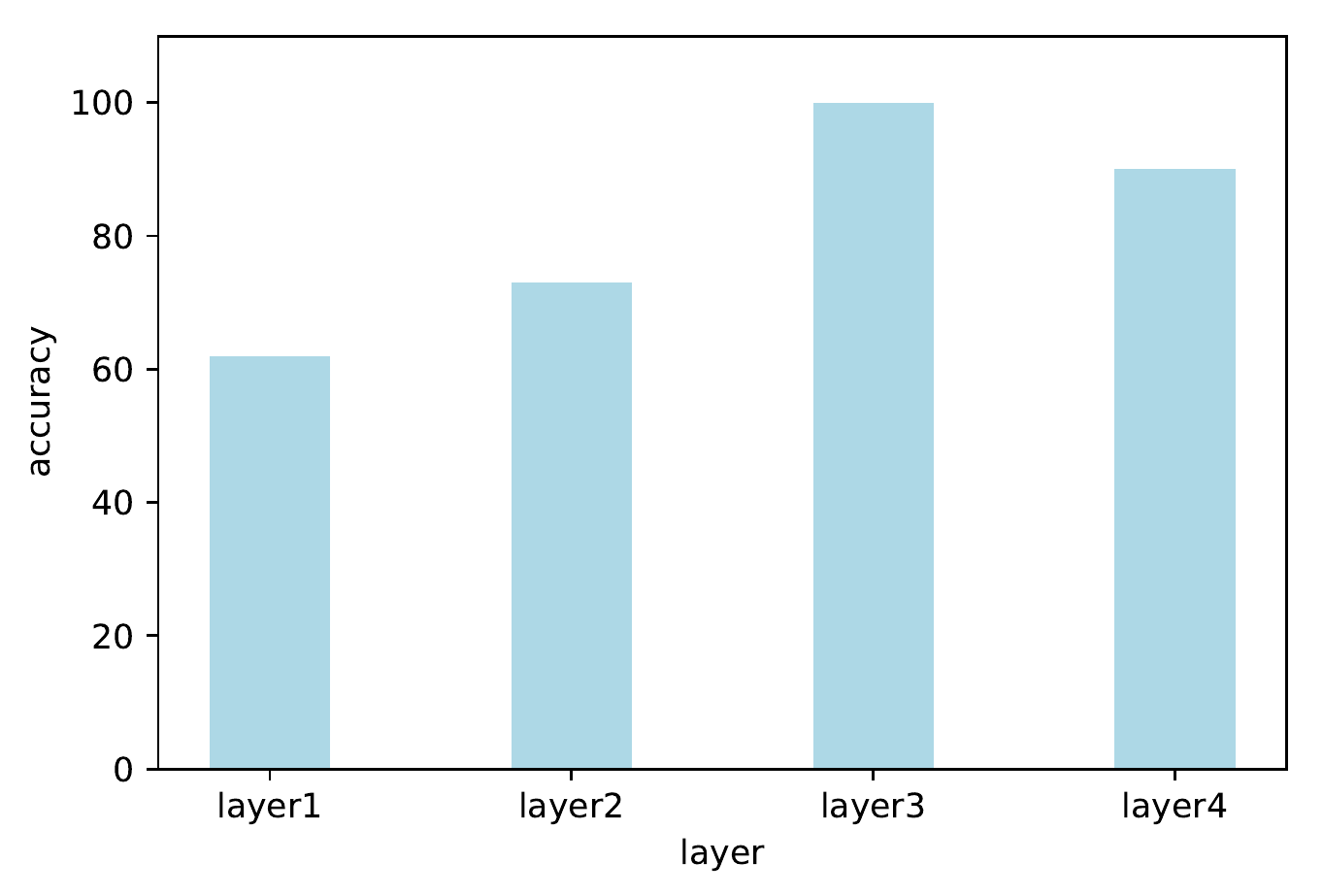}}
\setlength{\belowcaptionskip}{-30pt}
\caption{Number of Concept Classifiers whose accuracy was above 0.75.  Once again, layer 3 outperforms the other layers.}
\label{fig:num_clfs_per_layer}
\end{center}
\end{figure}

\subsection{Concept Classifier Prediction Performance}
Concept classifier accuracy varied across the different layers of Resnet. Figure~\ref{fig:avg_clf_accuracy_per_layer} shows that the average classifier accuracy was the highest for the third layer (0.844).  Unsurprisingly, as Figure~\ref{fig:num_clfs_per_layer} shows, it also produced the highest number of classifiers whose accuracy was above $\lambda$. 
\begin{figure}[ht]
\begin{center}
\centerline{\includegraphics[width=0.95\columnwidth]{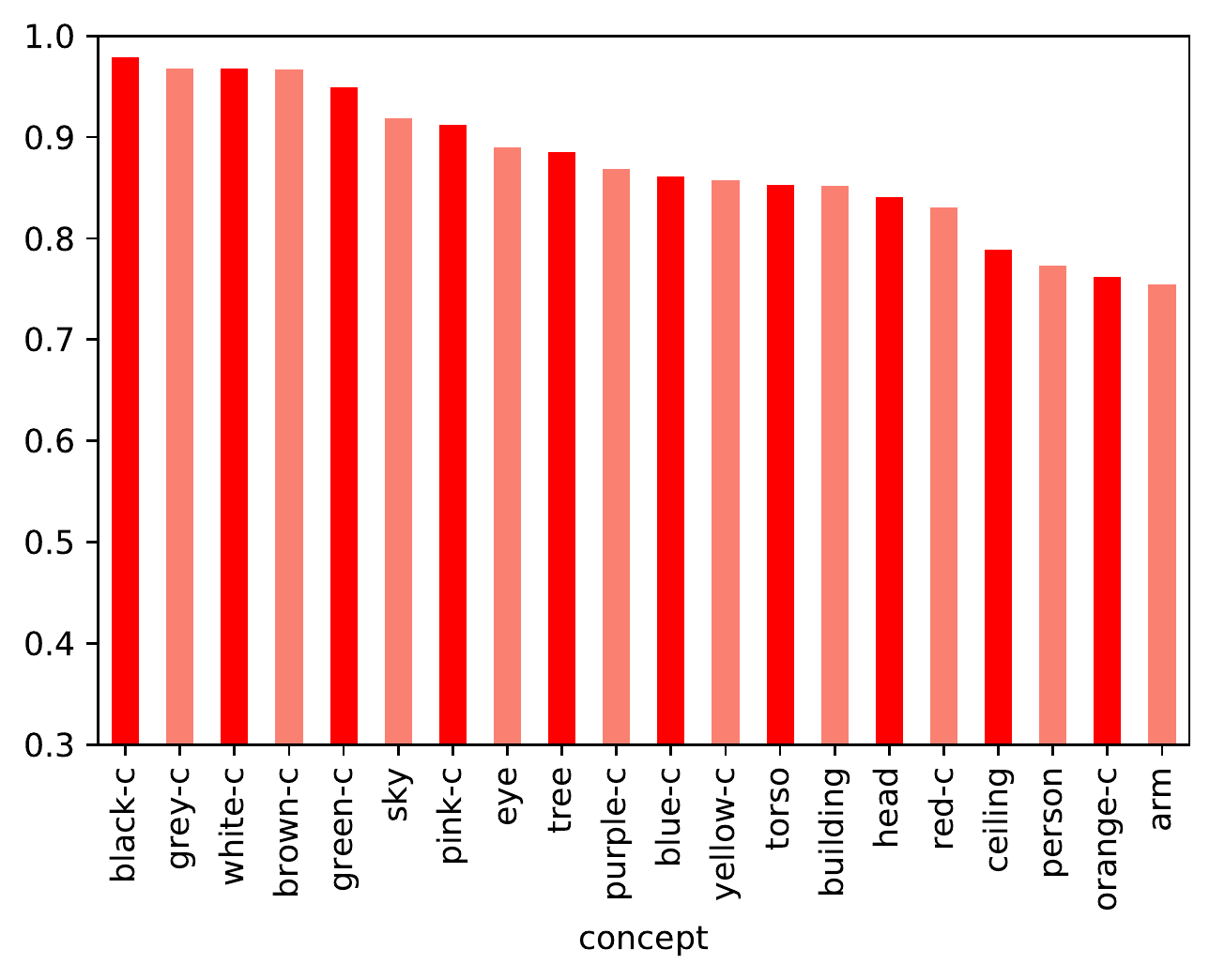}}
\setlength{\belowcaptionskip}{-30pt}
\caption{Concept classifier accuracy for the top 20 concepts trained using layer 3 of Resnet50. The top concepts are colors, i.e., black, grey, white, etc.}
\label{fig:accuracy_vs_concept}
\end{center}
\end{figure}

In general, concepts with more labeled examples from BRODEN achieved a higher accuracy.  Regardless of which layer the activations were extracted from, concepts of the color categories generally achieved the highest accuracy.  Figure~\ref{fig:accuracy_vs_concept} shows the distribution of accuracy scores for the third layer of Resnet for the top 20 scoring concept classifiers.  The top 5 are colors, but other general concepts such as \textit{sky}, \textit{building}, and \textit{head} also scored well.

\begin{figure*}[t]
\begin{subfigure}{0.49\textwidth}
\centering
\includegraphics[scale=0.55]{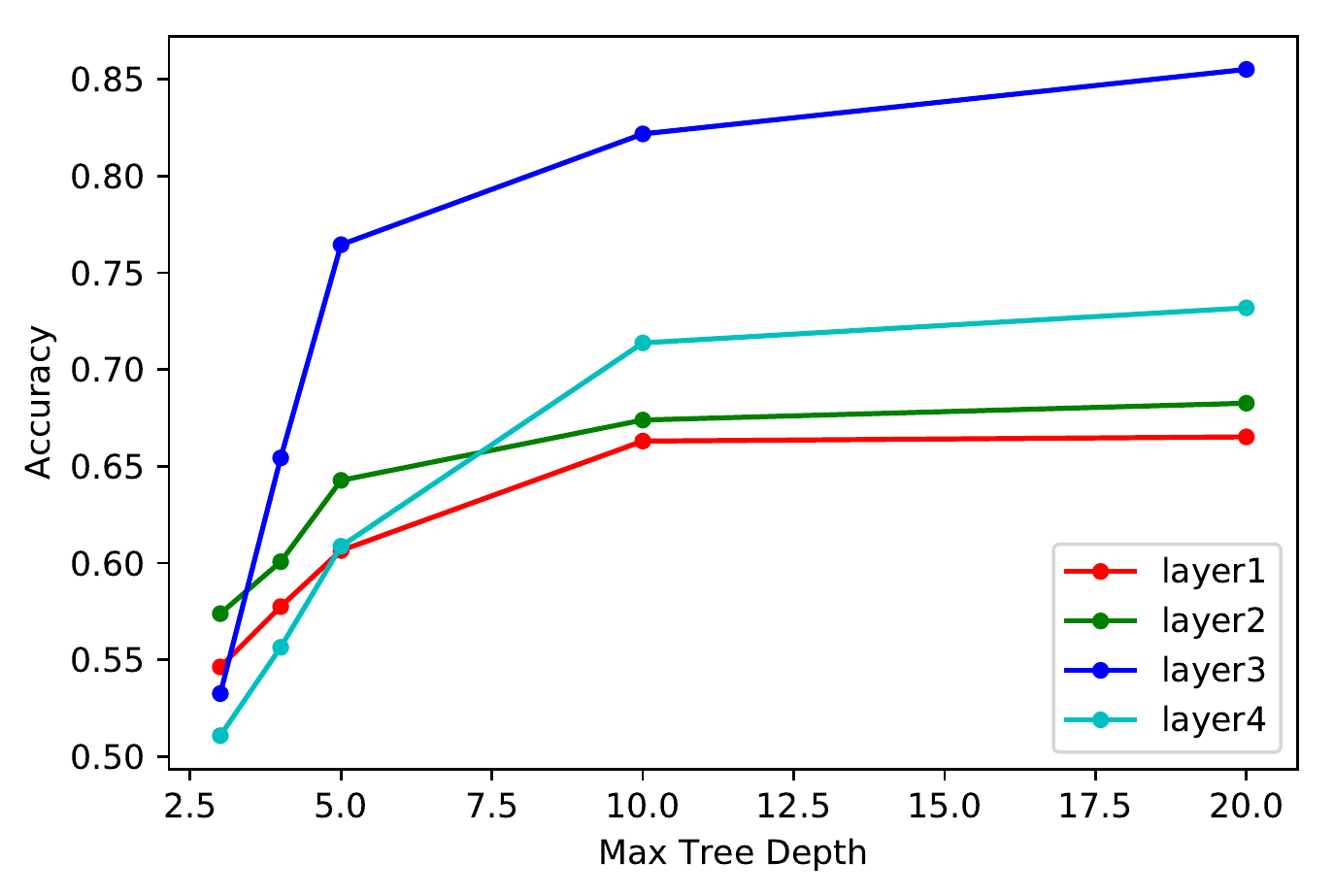}
\caption{Natural Images dataset.}
\label{fig:natural_images}
\end{subfigure} %
\hspace{0.2in}
\begin{subfigure}{0.49\textwidth}
\centering
\includegraphics[scale=0.55]{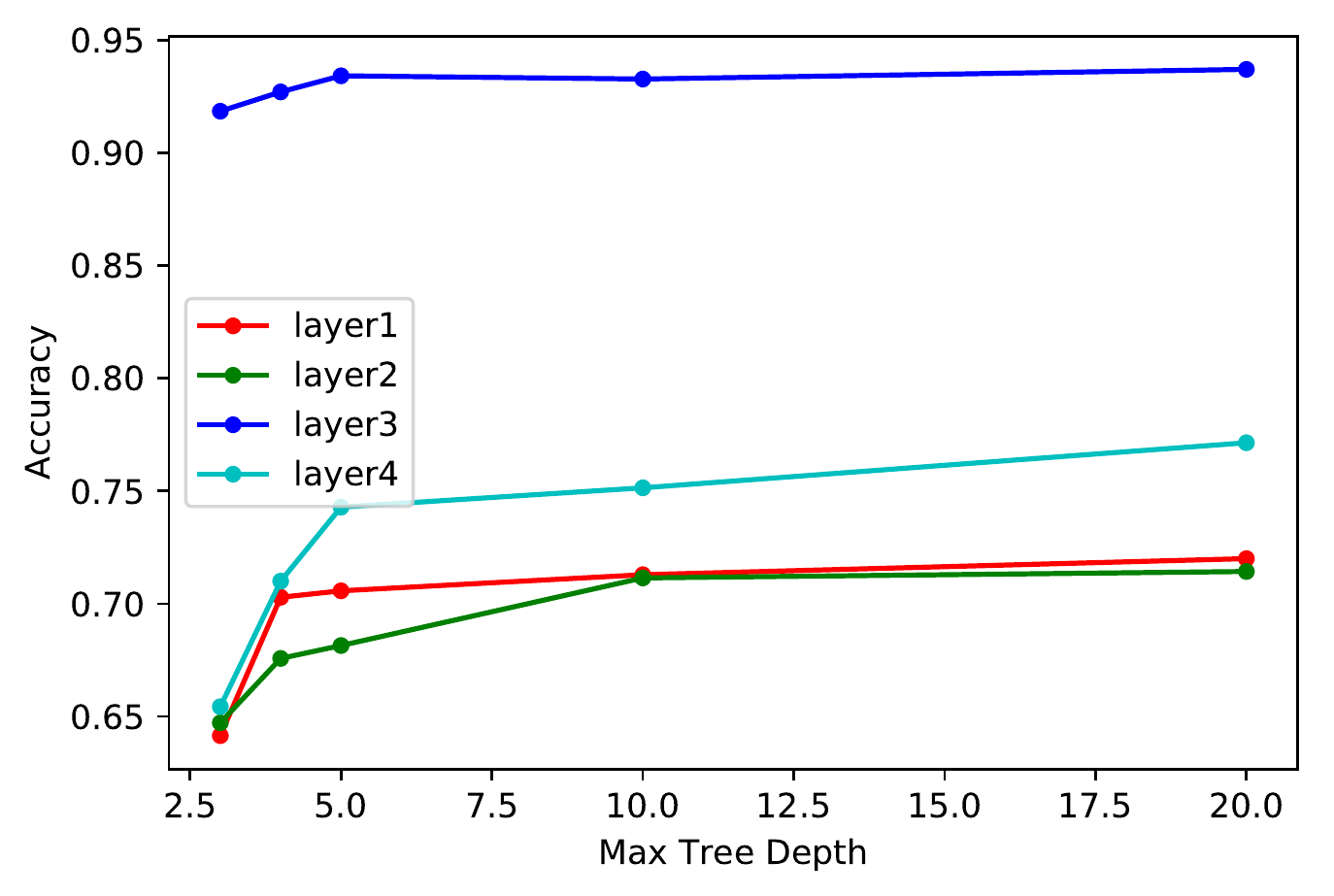}
\caption{Mini
Natural Images dataset. For this dataset, the accuracy maxes out
quickly as max tree depth increases}
\label{fig:mini_natural_images}
\end{subfigure}
\caption{Decision tree accuracy vs. max. tree depth for respective datasets}
\end{figure*}

\subsection{Decision Tree Performance}
Figures~\ref{fig:natural_images} and~\ref{fig:mini_natural_images} show how decision tree accuracy responds to changes in max. tree depth for each layer of Resnet50.  Accuracy improves greatly at first as the max. tree depth is increased, but quickly begins to flatten out as depth increases beyond 10 levels. 

Both plots demonstrate how layer 3 of Resnet50 has the best performance in terms of decision tree accuracy.  This is especially evident in the decision tree accuracies for the Mini Natural Images dataset with the best decision trees reaching accuracy scores in the low 90s.

\subsection{Interpretability}
While training a decision tree with increased depth leads to higher prediction accuracy w.r.t the original model, it also leads to a less interpretable result as the number of nodes increases exponentially with depth.  Thus a shallower decision tree with similar is preferred.

\begin{figure}[h]
\begin{center}
\vspace{-0.27in}
\centerline{\includegraphics[width=0.9\columnwidth]{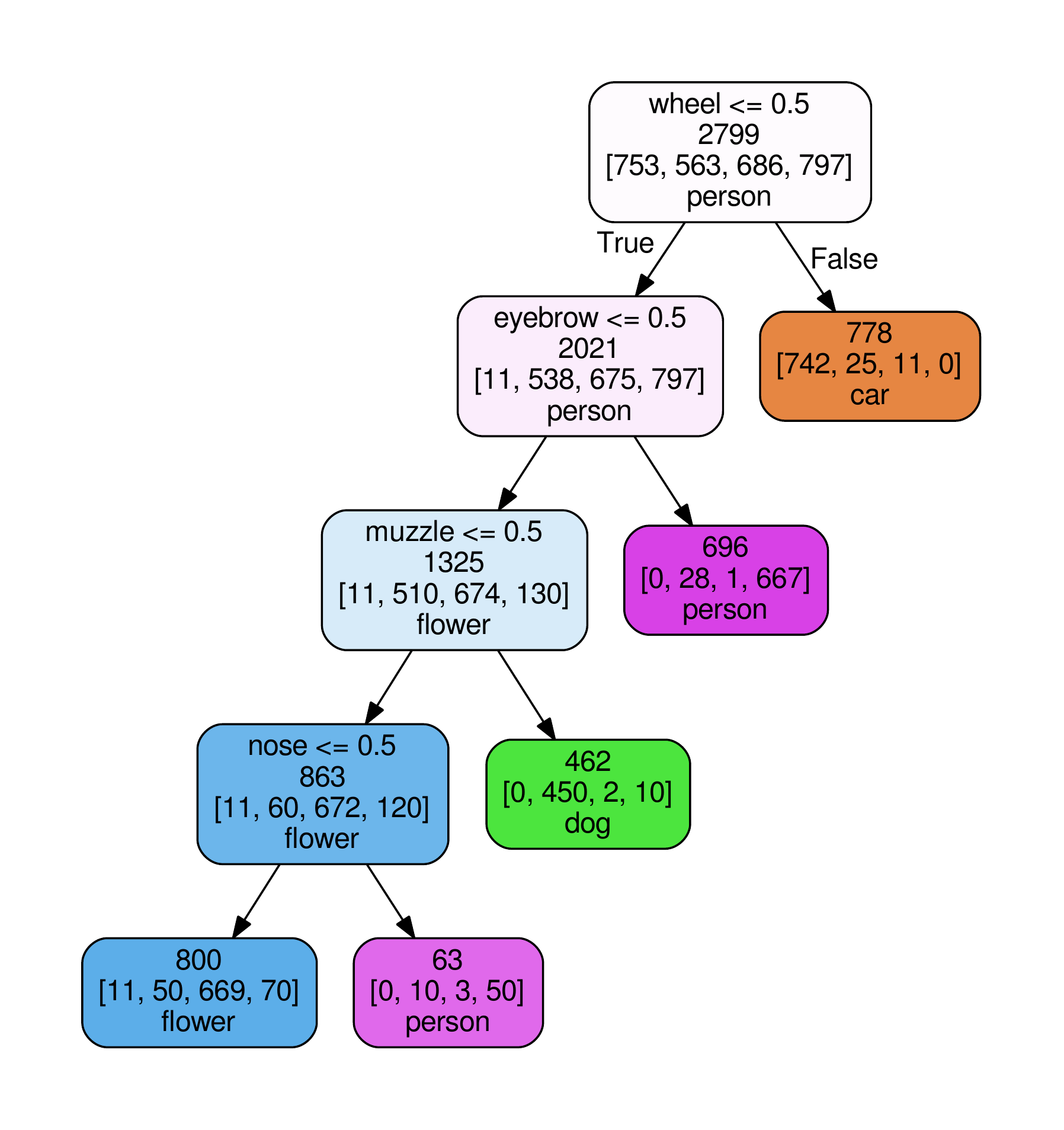}}
\vspace{-.12in}
\caption{Shallow decision tree of max. depth 5 and min. sample size of 20 trained on layer 3 ({conv4\_x}) of Resnet50.  The left branch indicates that the concept is not present, while the right branch indicates that the concept is present.  
The tree gives insight into how the model might be making a prediction based on concepts for the Mini Natural Images dataset. The decision tree provides a natural and logical explanation for each path, i.e. the presence of 'wheel' indicates 'car', the presence of 'eyebrow' but not 'wheel' indicates a 'person', etc.}
\vspace{-.12in}
\label{fig:decision_tree}
\end{center}
\end{figure}

Figure~\ref{fig:decision_tree} shows how a shallow decision tree can be trained to match the representation of Resnet50 trained on Mini Natural Images.  The learned decision tree is able to provide reasonable explanations that apply to all input images for each class in Mini Natural Images.  At the same time this specific tree was able achieve a relatively high accuracy (0.9134).  Figure~\ref{fig:mini_natural_images} shows that this tree also achieved an accuracy very similar to deeper trees trained on the same layer.

\section{Conclusion and Discussion}
A shallow decision tree with high accuracy w.r.t. the representation of the original model being probed gives insight into how the model might be reasoning about human-understandable concepts while making its prediction

Using this interpretation, one can infer which concepts are significant to the model based on which nodes are included in the decision tree.  

It also provides an interpretable alternative to the CNN while still maintaining competitive performance.

Because this method is portable, it can be applied to any CNN and can be used to extract domain-specific for any given problem.

The decision tree can also be a useful tool for detecting bias learned by the CNN if it is found that a certain concept is a discriminating feature that should not necessarily be informing the overall decision.

Additionally, extracting concept predictions and training the decision tree is generally a fast process and is only hamstrung by the speed in which inferences can be run through the model.  The time it takes to to get predictions from the concept classifiers and fit the decision tree consistently remains under a few seconds.  This allows for efficient tuning of hyperparameters to create a tree that is optimal in terms of interpretability and accuracy.

Future work along the route of concept-based decision trees could include a method that can extract concept classifiers from each layer of the network together and analyzing which types of concepts are best classified at each layer.  The highest performing concept classifiers from each each layer could also be combined together in an ensemble to provide a more holistic view of how concepts interact through the entire network.  

Instead of using binary classifiers to detect concepts from hidden layer activations, regressors could be used to extract more fine-grained concept data.  This could potentially lead to higher accuracy and more interpretable results; even for larger image classification problems.

\clearpage
\bibliography{example_paper}
\bibliographystyle{icml2019}
\end{document}